%% file: DiurnalCyclePersistenceFelix_arxiv.tex
\documentclass[dvipsnames]{article}
\usepackage[margin=0.75in]{geometry}
\usepackage{graphicx,amsmath,upgreek,bm}
\usepackage{wrapfig}
\usepackage[rgb, table]{xcolor}
\usepackage{cite}
\usepackage[toc,page]{appendix}
\usepackage{algorithm}
\usepackage{rotating}
\usepackage{hyperref}
\hypersetup{
   colorlinks=true,
   linkcolor=blue,
   filecolor=magenta,
   urlcolor=blue,
   citecolor=blue,
	}
\usepackage{caption}
\usepackage{subcaption}
\usepackage{amsmath}
\usepackage{gensymb}
\usepackage{mathtools}
\usepackage{amssymb}
\usepackage{appendix}
\usepackage{booktabs}
\usepackage{graphicx}
\usepackage{textcomp}
\usepackage{xcolor}
\usepackage{float}
\usepackage{multicol}
\usepackage{makecell}
\usepackage{colortbl}
\usepackage{stfloats}

\usepackage{Liz}

\graphicspath{{Figures/}{Figures}}

\title{Using persistent homology to quantify a diurnal cycle in hurricanes}

\usepackage{authblk}

\author[1]{Sarah Tymochko}
\author[1,2]{Elizabeth Munch}
\author[3,4]{Jason Dunion}
\author[5]{Kristen Corbosiero}
\author[5]{Ryan Torn}
\affil[1]{\small{Michigan State University, Dept. of Computational Mathematics, Science and Engineering}}
\affil[2]{Michigan State University, Dept. of Mathematics}
\affil[3]{Cooperative Institute for Marine and Atmospheric Studies, University of Miami}
\affil[4]{Hurricane Research Division, NOAA/Atlantic Oceanographic and Meteorological Laboratory}
\affil[5]{University at Albany - SUNY Albany, Dept. of Atmospheric and Environmental Sciences}

\date{}

\begin{document}

\twocolumn[
\begin{@twocolumnfalse}

  \maketitle
  \begin{abstract}
  The diurnal cycle of tropical cyclones (TCs) is a daily cycle in clouds that appears in satellite images and may have implications for TC structure and intensity. The diurnal pattern can be seen in infrared (IR) satellite imagery as cyclical pulses in the cloud field that propagate radially outward from the center of nearly all Atlantic-basin TCs. These diurnal pulses, a distinguishing characteristic of this diurnal cycle, begin forming in the storm's inner core near sunset each day, appearing as a region of cooling cloud-top temperatures. The area of cooling takes on a ring-like appearance as cloud-top warming occurs on its inside edge and the cooling moves away from the storm overnight, reaching several hundred kilometers from the circulation center by the following afternoon. The state-of-the-art TC diurnal cycle measurement in IR satellite imagery has a limited ability to analyze the behavior beyond qualitative observations. We present a method for quantifying the TC diurnal cycle using one-dimensional persistent homology, a tool from Topological Data Analysis, by tracking maximum persistence and quantifying the cycle using the discrete Fourier transform. Using Geostationary Operational Environmental Satellite IR imagery from Hurricanes Felix and Ivan, our method is able to detect an approximate daily cycle.
  \end{abstract}


  \vspace{1cm}

\end{@twocolumnfalse}
]

\input{sec-Intro}

\input{sec-Bkgd-Atmos}

\input{sec-Bkgd-Math}

\input{sec-Method}

\input{sec-Results}

\input{sec-Discussion}

\section*{Acknowledgments}
This manuscript is under consideration at Pattern Recognition Letters.
The authors would like to thank William Dong, who was instrumental in creating code for the first iteration of this project.
The authors also thank the two anonymous reviewers whose suggestions improved the quality of the paper.
ST and EM were supported in part by NSF grants DMS-1800446 and CMMI-1800466; EM was also supported in part by NSF CCF-1907591.
JD was supported in part by the Office of Naval Research Program Element (PE) \linebreak 0601153N Grant N000141410132 and by NOAA Grant NA14OAR4830172 from the Unmanned Aircraft Systems (UAS) Program.
KC was supported by NSF Grant AGS-1636799.

\bibliographystyle{abbrv}
\bibliography{HurricaneFelix}

\end{document}

%% file: sec-Intro.tex
\section{Introduction}
\label{sec:Intro}

The field of atmospheric science has numerous observation platforms that provide high space and time resolution data, but has yet to find methods which can quantify the intuitive patterns explicitly.
Meanwhile, the young field of Topological Data Analysis (TDA) encompasses methods for quantifying exactly these sorts of structural intuitions seen by atmospheric scientists.
This paper merges these two fields by using persistent homology, a now well-established tool in TDA, to quantify a diurnal cycle observed in a hurricane using Geostationary Operational Environmental Satellite (GOES) infrared (IR) satellite data.

Persistent homology, and more generally TDA methods, has found significant success in rather disparate applications by finding structure in data and using this insight to answer questions from the domain of interest.
For instance, Giusti et al.~used the homology of random simplicial complexes to investigate the geometric organization of neurons in rat brains (\cite{Giusti2015}).
Persistent homology has also been used to understand periodicity in time series arising from biological (\cite{Deckard2013, Perea2015}) and engineering applications (\cite{Khasawneh2015}), as well as for image analysis (\cite{Carlsson2008,RobinsWoodSheppard2011,Robins2016,Saadatfar2017}).

This paper presents an application of the use of both time series and image analysis using TDA to study a daily cycle in hurricanes that is of interest in the field of atmospheric science.
The diurnal cycle of tropical cyclones (TCs) has been described in previous studies (\cite{Dunion2014, dunion2019tropical, Ditchek2019a,Ditchek2019b, kossin2002daily, leppert2016tropical, navarro2016idealized, o2017accessible, steranka1984diurnal, tang2016impacts}) that provide evidence of the regularity of this cycle as well as its potential impacts.
This diurnal pattern can be seen in GOES IR imagery as cyclical pulses in the cloud field that propagate radially outward from  TCs at speeds of 5-10 m s$^{-1}$ (\cite{Dunion2014, dunion2019tropical,Ditchek2019a}).
These diurnal pulses, a distinguishing characteristic of the TC diurnal cycle, begin forming in the TC's core near the time of sunset each day and appear as a region of cooling cloud-top temperatures.
The area of cooling then takes on a ring-like appearance as marked cloud-top warming occurs on its inside edge and it moves away from the storm overnight, reaching several hundred kilometers from the TC center by the following afternoon.
Observations and numerical model simulations indicate that TC diurnal pulses propagate through a deep layer of the TC environment, suggesting that they may have implications for TC structure and intensity (\cite{Ditchek2019a,Ditchek2019b, Dunion2014, dunion2019tropical,  leppert2016tropical}).

The current state of the art TC diurnal cycle measurement has a limited ability to analyze the behavior beyond qualitative observations. This paper presents a more advanced mathematical method for quantifying the TC diurnal cycle using tools from TDA, namely one-dimensional persistent homology to analyze the holes in a space. This research aims to detect the presence of the diurnal cycle in GOES IR satellite imagery and to track the changes through a time series.

In this paper, we present a method of automatically detecting and quantifying periodic circular structure in satellite imagery as well as the results of our method applied to GOES IR hurricane imagery for Hurricane Felix in 2007 and Hurricane Ivan in 2004.
The naive combination of persistent homology with the IR imagery did not show a recurring pattern due to drastically variable values in IR brightness temperature data.
Despite this, looking at the data, there is a clear circular feature that is visible in the imagery.
Thus we develop a more sophisticated method using tools including the distance transform and one-dimensional persistent homology to detect the TC diurnal cycle quantitatively using maximum persistence.
Using our method, we are able to detect this 24-hour cycle in both hurricanes automatically, improving upon the existing qualitative methods.

%% file: sec-Bkgd-Atmos.tex
\section{Tropical Cyclone Background}
\label{sec:HurricaneBkgd}

Previous research has documented a clear diurnal cycle of cloudiness and rainfall in TCs: enhanced convection (i.e., thunderstorms) occurs overnight, precipitation peaks near sunrise, and upper-level cloudiness (i.e., the cirrus canopy) expands radially outward throughout the day, reaching its maximum areal coverage in the early evening hours (\cite{Ditchek2019a,Ditchek2019b, Dunion2014, dunion2019tropical, kossin2002daily, leppert2016tropical, navarro2016idealized, o2017accessible,  steranka1984diurnal, tang2016impacts}).
To quantify the expansion and contraction of the cirrus canopy, Dunion et al. used GOES satellite IR imagery to examine the six-hour cloud-top temperature differences of major hurricanes in the Atlantic basin from 2001 to 2010 (\cite{Dunion2014}).
They found that an area of colder cloud tops propagated outward around 5-10 m s$^{-1}$ over the course of the day, with warming temperatures on its inner edge.
More recently, in (\cite{Ditchek2019a}), Ditchek et al. expanded Dunion et al.'s work to include all tropical cyclones in the Atlantic basin from 1982 to 2017 and found that the diurnal pulse is nearly ubiquitous, with 88\% of TC days featuring an outwardly propagating pulse.

Despite the consistent signature and documentation of this diurnal cloud signature, open questions remain as to how the diurnal cycle is linked to inner-core convective processes and whether it is a column-deep phenomenon or mainly tied to upper-level TC cloud dynamics related to incoming solar radiation (\cite{Ditchek2019a, navarro2016idealized, o2017accessible, tang2016impacts}).
Investigating these questions is relevant to TC forecasting as the diurnal cycle of clouds and rainfall has implications for forecasting storm structure and intensity, as evidenced by the diurnal cycle in objective measures of TC intensity and the extent of the 50-kt wind radius documented by Dunion et al.
Additionally, and especially relevant to the current work, most of the papers above have identified the pulse using subjective measures of cloud-top temperature change and timing (\cite{Ditchek2019a, Dunion2014}).
The current work seeks to quantify the pulse to determine its true periodicity using persistent homology, a topological tool that is particularly effective at capturing the type of patterns visible in the pulse.

%% file: sec-Bkgd-Math.tex
\section{Math Background}
\label{sec:MathBkgd}

Persistent homology is a tool from the field of TDA which measures structure in data.
This data can start in many forms, including as point clouds or, as in the case of this work, as a function on a domain.
In this section, we will briefly review the necessary background to understand cubical complexes and persistent homology, and refer the interested reader to \cite{Edelsbrunner2010, Hatcher, KaczynskiMischaikowMrozek2006, Munch2017} for a more complete introduction.

\subsection{Cubical complexes}
\label{ssec:CubComplex}

\begin{figure}
\centering
 \includegraphics[width=.2\textwidth]{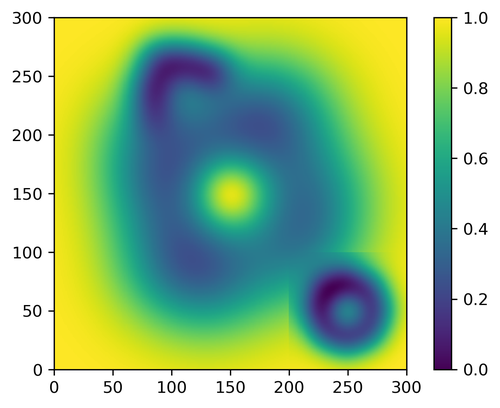}
 \includegraphics[width=.17\textwidth]{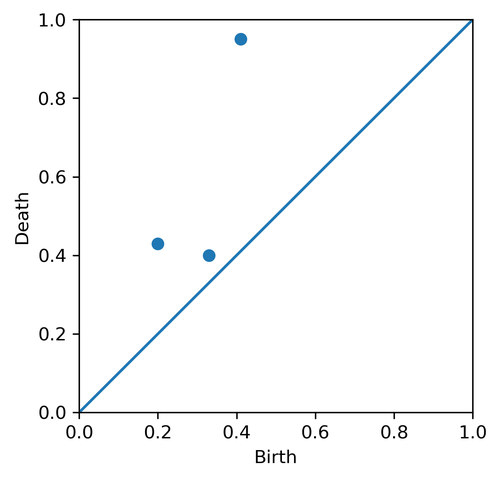} \\
  \includegraphics[width=.14\textwidth]{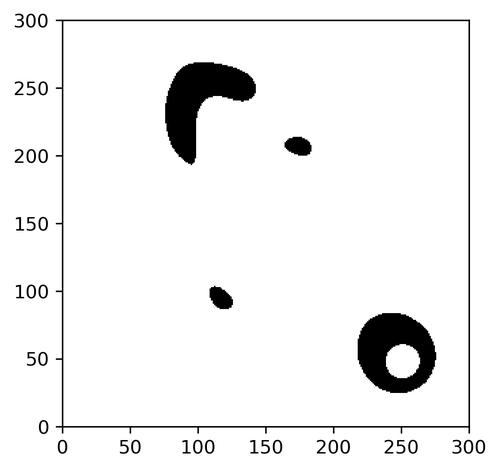}
 \includegraphics[width=.14\textwidth]{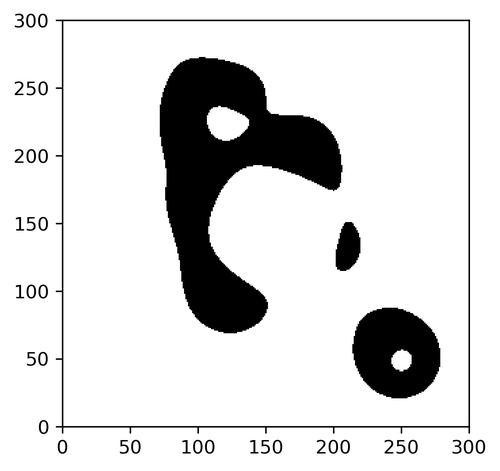}
 \includegraphics[width=.14\textwidth]{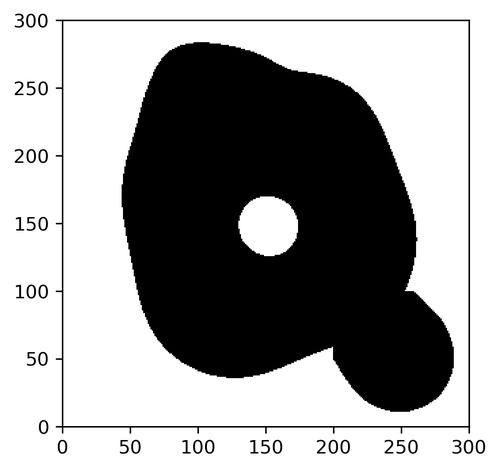}

 \caption{An example matrix, $M$ (top left) and corresponding persistence diagram (top right). Second row: The black portions are sublevel sets, $M_r$, where $r = 0.25, 0.35,$ and $0.7$. The existence of a point far from the diagonal in the persistence diagram shows that there is a prominent circular structure; while the other points are caused by the smaller circles. }
 \label{fig:PersForMatrices}
\end{figure}

In this section, we largely follow Chapter 2 of \cite{KaczynskiMischaikowMrozek2006} with the caveat for the informed reader that because we use homology with $\Z_2$ coefficients, we can be lazy about orientations of cubes.
In addition, our data consists of 2D images, so we need only define cubes up to dimension 2.

An elementary interval is a closed interval $I \subset \R$ of the form $[\ell,\ell+1]$ or $[\ell]$ for $\ell \in \Z$, which are called nondegenerate and degenerate respectively.
An elementary cube $Q \in \R^2$ is a product of elementary intervals $Q = I_1 \times I_2$.
The dimension of $Q$, $\dim(Q)$, is the number of nondegenerate components of $Q$.
Note that 0-dimensional cubes are just vertices at the points on the lattice $\Z \times \Z$ in $\R^2$, 1-dimensional cubes are edges connecting these vertices, and 2-dimensional cubes are squares.
Let $\KK$ denote the set of all elementary cubes in $\R^2$ and $\KK_d\subset \KK$ the set of $d$-dimensional cubes.
A set $X \subset \R^2$ is cubical if it can be written as a finite union of elementary cubes.
Then we denote the associated cubical complex as $\KK(X) = \{Q \in \KK \mid Q \subset X\}$, with the $d$-dimensional subset denoted $\KK_d(X) = \{Q \in \KK(X) \mid \dim(Q) = d\}$.
If $Q \subseteq P$, then we say $Q$ is a face of $P$, denoted $Q \leq P$.
If $Q \subsetneq P$, then $Q$ is a proper face of $P$, denoted $Q < P$, and is additionally a primary face of $P$ if $\dim(Q) = \dim(P)-1$.

A greyscale image, or more generally an $m \times n$ matrix, can be viewed as a function $M: D \to \R$ where
	$D  = \{ (i,j) \mid 0 \leq i < m, 0\leq j < n\}$.
We will model this as a function defined on a particularly simple cubical set $K = \KK([0,m] \times [0,n])$. %
For simplicity, we denote by $s_{i,j}$ the square $[i,i+1] \times [j,j+1]$.
So, given a matrix $M$, we equivalently think of this data as a function $M: K \to \R$ where we set $M(s_{i,j})$ equal to the matrix entry $M_{i,j}$ and set $M(P) = \min_{s_{i,j}> P} M(s_{i,j})$ for all lower dimensional cubes $P$.
Note that we will abuse notation and use $M$ to denote both the original matrix and representation as a function with domain $K$.

\subsection{Homology}

Homology (\cite{Hatcher}) is a standard tool in algebraic topology which provides a vector space\footnote{Normally a group, however, we are working with field coefficients.} $H_k(X)$ for each dimension $k=0,1,2,\dots$ for a given topological space $X$.
The different dimensions measure different properties of the space.
In particular, for this work we are interested in $1$-dimensional homology; i.e. when $k=1$.
The 1-dimensional homology group measures the number of loops in the space; equivalently, we can think of this as the number of holes in the space.
In particular, if we look at the black region in each of the examples in Fig.~\ref{fig:PersForMatrices}, the rank of the first homology for each is (1,2,1).

The exact definition of homology is as follows.
For any cubical set $L$ (which for the purposes of this discussion will always be a subset of $K$), we have sets giving the cubes of different dimensions: $\KK_i(L)$ for $i = 0,1,2$.
An $i$-chain is a formal linear combination of $i$-simplices in $L$,
$c = \sum_{Q_j \in \KK_i(L)} a_j Q_j,$
with coefficients $a_j \in \Z_2$.
We can of course add these objects by setting
$\left( \sum a_jQ_j\right) + \left( \sum b_jQ_j\right) =  \sum (a_j+b_j)Q_j$ and multiply by a constant.
Thus, the collection of all $i$-chains forms a vector space $C_i(L)$.

We  define a linear transformation $\delta_i: C_i(L) \to C_{i-1}(L)$ called the boundary map, by setting $\delta_i(Q) = \sum P$ where the sum%
\footnote{Again, notice that because we are working with $\Z_2$ coefficients, the book-keeping normally needed for orientation is unnecessary.}
is over the primary faces $P < Q$.
The kernel of $\delta_1$, $\Ker(\delta_i)$, (that is, the set of elements of $C_1(L)$ which map to 0) is generated by closed loops in $L$.
The image of $\delta_2$, $\Im(\delta_2)$, is generated by boundaries of 2-cells.
Then the $1$-dimensional homology group is defined to be $H_1(K) = \Ker(\delta_1)/\Im(\delta_{2})$.
An element of this group $\gamma \in H_1(L)$, represents an equivalence class of loops which can differ by collections of 2-cells.

\subsection{Persistent Homology}
\label{ssec:PersHomology}
For a \textit{static} space $L$, $H_1(L)$ measures information about the number of loops.
Persistent homology takes as input a \textit{changing} topological space, and summarizes the information about how the homology changes.

Let an $m \times n$ $\R$-valued matrix $M$ be given.
Fix a function value $r \in \R$ and let $M_r = f\inv(-\infty, r]$.
That is, $M_r$ is the subset of squares in $K_{m \times n}$  which have value at most $r$ in the matrix, along with all edges and vertices which are faces of any included square.
$M_r$ is often called a sublevel set of $M$.
See Fig.~\ref{fig:PersForMatrices} for $M_r$ regions corresponding to an example image.

These spaces have the property that $M_r \subseteq M_s$ for $r\leq s$, thus we can consider the sequence
$
M_{r_1} \subseteq M_{r_2} \subseteq \cdots \subseteq M_{r_k}
$
for any set of numbers $r_1 < r_2< \cdots r_k$.
This sequence of spaces is called a filtration.
For each of the spaces, we can compute the homology group $H_p(M_{r_i})$.
The inclusion maps give rise to linear maps
$
H_p(M_{r_1}) \rightarrow H_p(M_{r_2}) \rightarrow \cdots \rightarrow H_p(M_{r_k}).
$
It is these maps that we study to understand how the space changes.
In particular, when we are focused on 1-dimensional homology ($k=1$) as in this study, a loop is represented by an element $\gamma \in H_1(M_{r_i})$.
We say that this loop is born at $r_i$ if it is not in the image from the previous space; that is, $\gamma \not \in \Im(H_1(M_{i-1}) \to H_1(M_i))$.
This same loop dies at $r_j$ if it merges with this image in $M_{r_j}$; that is, $\gamma \in \Im(H_1(M_{i-1}) \to H_1(M_j))$ where we abuse notation by using $\gamma$ to both refer to the class in $H_1(M_{r_i})$ and the image of this class under the sequence of maps in $H_1(M_{r_j})$.
We refer to $r_j-r_i$ as the lifetime of the class.

A persistence diagram, as seen in the right of Fig.~\ref{fig:PersForMatrices}, is a collection of points where for each class which is born at $r_i$ and dies at $r_j$ is represented by a point at $(r_i,r_j)$.
The intuition is that a class which has a long lifetime is far from the diagonal while a class with a short lifetime is close.
In many cases, a long lifetime loop implies that there is some sort of inherent topological feature being found, and thus that this point far from the diagonal is important, while short lifetime loops are likely caused by topological noise due to sampling or other errors in the system.
In the example of Fig.~\ref{fig:PersForMatrices}, there is a prominent off-diagonal point which shows that the function defined by the matrix has a circular feature.
Thus, a common measure for looking at the persistence diagram when investigating a single, circular structure is the maximum persistence, defined as
\begin{equation}
\mathrm{MaxPers}(D) = \max_{(r_i,r_j) \in D} r_j-r_i
\label{eqn:maxpers}
\end{equation}
for a given persistence diagram, $D$.

%% file: sec-Method.tex
\section{Methods}

\begin{figure}[tb]
	\centering
	\includegraphics[width = .21\textwidth]{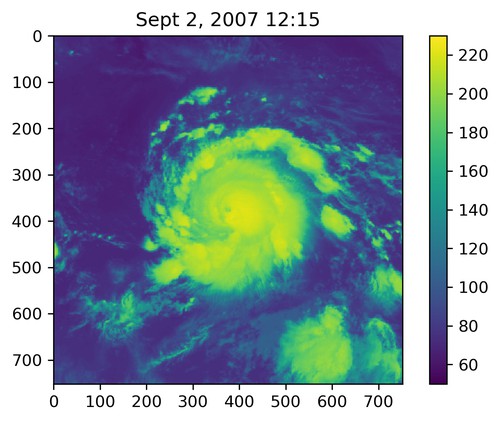}
	\includegraphics[width = .21\textwidth]{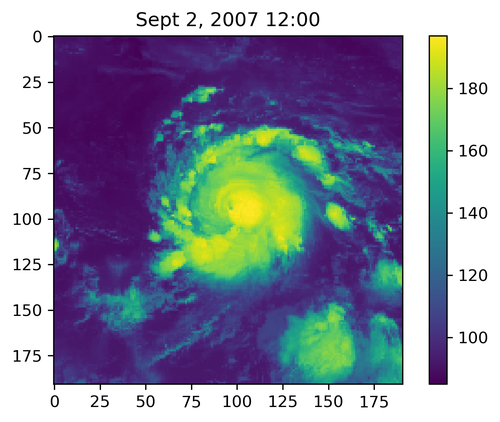}
	\caption{Original satellite imagery from Felix GOES-12 data set (left) and Felix GridSat-GOES data set (right) at approximately the same time.}%
	\label{fig:SatImages}
\end{figure}

\subsection{Data preparation}
\label{ssec:DataPrep}

The data was given in the form of storm-centered GOES IR (10.7-$\mu$m) satellite imagery.
The 10.7-$\mu$m long channel detects IR energy emitted from the Earth and is not strongly affected by atmospheric water vapor.
Thus, this particular channel is useful for detecting clouds at all times of the day and night and is ideal for tracing the diurnal evolution of the TC cloud fields.
For our analysis, we used two different types of data sets.
These two data sets have the same native spatial resolution, but differ in temporal resolution.
The first type (hereafter the GOES-12 data sets), utilizes brightness temperatures derived directly from GOES-12 4-km IR satellite imagery and consists of data in hourly increments, with the exception of 0415 and 0515 UTC each day in Hurricane Felix, and 0445 and 0545 UTC each day in Hurricane Ivan (due to the GOES-12 satellite eclipse period).
Imagery was remapped such that each pixel has a spatial resolution of 2 km$^2$ and each image covers a total area of approximately 1500 km $\times$ 1500 km, represented as a 752 $\times$ 752 matrix.
This remapping was performed using the Man Computer Interactive Data Access System (McIDAS; \cite{lazzara1999man}) in order to generate storm-centered satellite images that were focused on the relevant TC environment.
The McIDAS 4 km to 2 km remapping procedure replicates the original 8-bit grayscale values such that none of the original pixel information is lost.
The second type is the GridSat-GOES (\cite{Knapp2018}) data set and consists of data in 3-hour increments with the exception of 0600 UTC each day.
Each pixel has a resolution of approximately 8 km and each image covers a total area of approximately 2400 km $\times$ 2400 km, represented by a 301 $\times$ 301 matrix.
This data is cropped to a 191 $\times$ 191 matrix to approximately match the area covered by the first set of data.
The cropped version covers a total area of approximately 1530 km $\times$ 1530 km.

The GridSat-GOES data set requires some additional processing.
The data is stored using a different format, using short numbers rather than floats, so the following equation is applied to the GridSat-GOES brightness temperatures to do the number type conversion:
$
[(\text{Original} \cdot 0.01 + 200.0)-22.858] / 0.919565.
$
Some images in the GridSat-GOES data set also contain missing values where the brightness temperature for certain pixels was not recorded and is instead assigned a fill value.
In order to prevent these values from impacting our results, we interpolate values for these pixels.
For a given pixel with a missing value, we compute the average value of a $5\times 5$ grid centered at the pixel, not including the pixels in this range that also have missing values.

For Hurricane Felix, we studied both types of data sets to test the flexibility of the method across spatial and temporal resolution.
The Felix GOES-12 data set spans 2 to 4 September 2007, while the Felix GridSat-GOES dataset spans spanning 31 August to 6 September 2007.
For Hurricane Ivan, we used the GOES-12 data set, which spans 30 August to 1 September 2004.

\subsection{Method of Detection \& Quantification}
\label{sec:method}

\begin{figure}[tb]
	\centering

	\includegraphics[width = .47\textwidth]{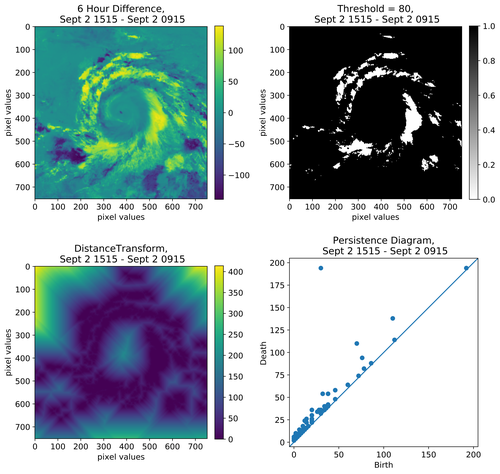}
	\caption{Top left: Example of 6 hour difference, $M(t)$, from the Felix GOES-12 data set; Top right: thresholded subset, $M(t)_{\mu}$ where $\mu=80$; Bottom left: distance transform function; Bottom right: corresponding persistence diagram. }
	\label{fig:StepsEx}
\end{figure}

We present a method of detecting and quantifying periodic circular structure representing the TC diurnal cycle in IR satellite imagery.
Our method combines existing methods from the fields of image processing, topological data analysis, and signal processing.

Initially, we have a time series of IR satellite images, represented as a matrix of pixel values $S(t)$ for time $t$.
The TC diurnal pulse is propagating outward through the day; thus, in order to see the movement and changes in the GOES satellite brightness temperature, we consider the difference in matrices six hours apart (\cite{Ditchek2019a,Dunion2014}).
For all times $t$, given the original brightness temperature image $S(t)$, we compute the six-hour differences, ${M(t) = S(t+6) - S(t)}$.
While circular features are visually prominent in the data, simply using persistence on the difference data did not show any relevant features.
This discrepancy is due to the extreme differences in the function values between the circular sections which prevents the sublevel sets from containing the full circular structure until very late in the filtration.
Thus, we are unable to detect the circular structure from $M(t)$ and we define a new function on the difference matrix using the following method.
Fix a threshold $\mu$ and let $M(t)_{\mu}$ be the subset of $M(t)$ which has function value less than $\mu$.
This method results in a binary matrix defined entry-wise, with $M(t)_{\mu}[i,j] = 1$ if $M(t)[i,j] < \mu$, and $M(t)_{\mu}[i,j] = 0$ otherwise.
We will address this choice of threshold in Sec.~\ref{ssec:threshold}; however, we will focus on the case $\mu = 80$ degrees for most of our analysis. Note that because $M(t)$ is a difference of two images, the threshold is not isolating all pixels above a certain temperature, but rather those pixels that increase in value by at least 80 degrees over the six hours.

After thresholding, we now have binary images; however, the persistent homology is uninteresting, as there are only two possible sublevel sets.
In order to create a greyscale image that maintains the visually apparent topological structure of the image, we apply the distance transform, a method from the field of image processing (\cite{Rosenfeld1966_distancetransform, Rosenfeld1968_distancetransform}), which gives a new matrix $D(t)$. %
The distance transform is calculated as follows: given any pixel $s_{i,j}$ in an image $M(t)$ represented as a matrix of pixels,
$D(t)_{i,j} = \min d(s_{i,j}, x)$
where $x$ is a 0-valued pixel and $d$ is any distance metric.
In this application, we specifically use the $L_\infty$ distance, also known as the chessboard distance; however, the distance transform is defined for any distance metric, the Euclidean distance transform being the most commonly used.
Given two pixels, $s_{i_1,j_1}, s_{i_2,j_2}$ the $L_\infty$ distance between them is calculated as
$d( s_{i_1,j_1}, s_{i_2,j_2} ) = \max  \{| i_2-i_1 |, |j_2-j_1|\}$.
This defines a distance on the pixels, which are the 2-cells in the cubical complex. The distance can be extended to the lower dimensional cells in the same manor as described in Sec. \ref{ssec:CubComplex}.

To calculate the distance transform, we use the python submodule \texttt{scipy.ndimage}, specifically the function \linebreak \texttt{distance\_transform\_cdt} with the chessboard metric.
\linebreak Therefore, each entry in $D(t)$ corresponds to the minimal distance to an entry where $M(t)[i,j] \geq \mu$.
The distance transform $D(t)$ is then scaled by the resolution for each data set in order to convert the distance units to kilometers instead of pixels.
For the GridSat-GOES data set, we scale by a factor of 8 km/pixel while for the GOES-12 data sets, we scale by a factor of 2 km/pixel.
We then compute sublevel set persistence on the function $D(t)$ using the \texttt{cubtop} method in Perseus (\cite{Mischaikow2013,Perseus}), which calculates persistent homology for cubical complexes using concepts from discrete Morse theory.
Note that Perseus requires an integer value filtration function on the cubical complex; thus, we chose to use the chessboard metric for the distance transform.
Figure~\ref{fig:StepsEx} shows an example of each step described so far.

For each six-hour difference in each data set, we apply the steps described above, then calculate maximum persistence as defined in Eqn.~\ref{eqn:maxpers}.
By plotting maximum persistence over time, we can see how the most prominent circular feature changes through the progression of the day and life of the TC.
This plot should show an oscillatory pattern, detecting the change in the diurnal cycle throughout the day.
In order to quantify this oscillatory pattern, we use the Fourier transform.
In general, the Fourier transform is a commonly used method for investigating periodicity of time series (\cite{Brigham1988}) by decomposing a wave into a sum of sinusoids with different frequencies.
Since we are working with discrete data, we will work with the discrete Fourier transform (DFT).
Let $T$ be the time between discrete samples, then let $t_k = kT$ where $k=1,\ldots, N-1$.
Then, the discrete Fourier transform is
$
F_n = \sum_{k=0}^{N-1} f(t_k) e^{-2\pi in k/N}.
$
This converts a function from the time domain to the frequency domain.
The power spectrum of $F_n$ can be estimated by calculating the square of the absolute value of the discrete Fourier transform, $|F_n|^2$.

Using the DFT, we calculate the most prominent frequency in the data in order to determine how often the cyclic behavior repeats.
Note, to use the discrete Fourier transform the time steps must be equal; however, because of the missing times in our data, this is not the case.
Therefore, we approximate the maximum persistence at these values by adding a point along the line between the times immediately before and after the missing time.
Additionally, we must truncate the maximum persistence to only include the days where we have the data for the entire day.
This means truncating the Felix GridSat-GOES data set to include only 1-4 September 2007, the Felix GOES-12 data set to include only 1-2 September 2007, and the Ivan GOES-12 data set to include only 30-31 August 2004. The discrete Fourier transform was calculated using the python submodule \texttt{numpy.fft}.

We first calculate the Fourier transform using the function \texttt{fft}, then calculate the frequency bins using \texttt{fftfreq}.
Using this information, we plot the approximate power spectrum for each data set.
Note, if working with real data as we are in this application, the power spectrum will be symmetric for positive and negative frequencies; therefore, we only need to look at the positive frequencies.
Picking the frequencies corresponding to the highest peaks in each power spectrum gives us the frequency of the most prominent periodic signal in the data.
Using this frequency, we can calculate the period of the oscillatory pattern and quantify the signal we are detecting.
To verify the detected signal matches the visual oscillatory pattern we see, we can reconstruct the sinusoid corresponding to the most prominent periodic signal using the inverse discrete Fourier transform, using \texttt{ifft}.
Then plotting this reconstructed sinusoid over the original data, we can see how closely this signal matches the patterns in the data.

%% file: sec-Results.tex
\section{Results}

\subsection{Experimental results}

\begin{figure}[tb]
	\centering
	\includegraphics[width = .37\textwidth]{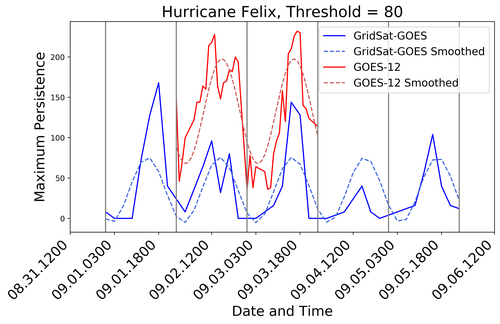} \\
	\includegraphics[width = .37\textwidth]{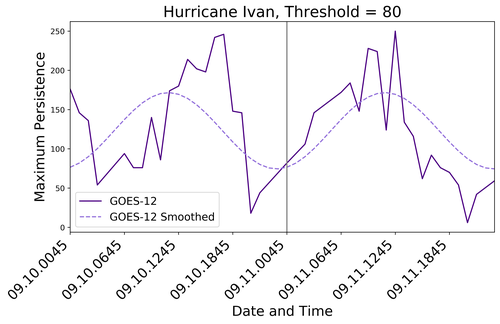}
	\caption{Maximum persistence plotted over time for all data sets using threshold $\mu=80$ in addition to the reconstructed versions, created using inverse Fourier transform. Gray vertical lines separate days according to UTC.}
	\label{fig:smoothed}
\end{figure}

\begin{figure}[tb]
	\centering
	\includegraphics[width = .21\textwidth]{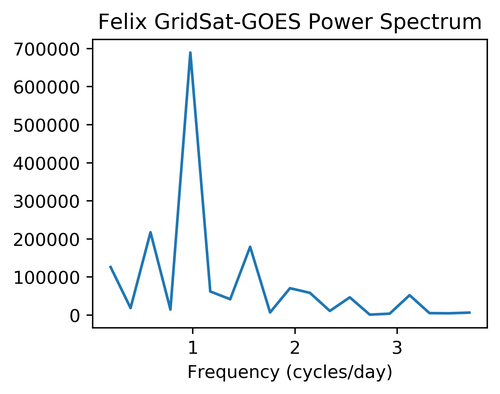}%
	\includegraphics[width = .22\textwidth]{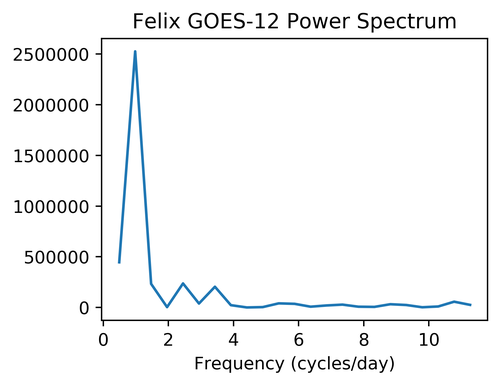} \\ %
	\includegraphics[width = .22\textwidth]{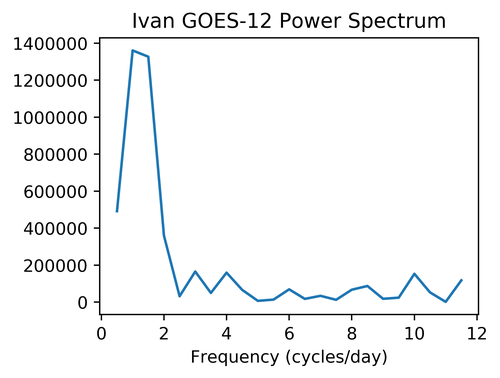}
	\caption{Power spectrum for each data set. The highest peak on the Felix GridSat-GOES power spectrum occurs at a frequency of approximately 0.976 cycles/day, the peak on the Felix GOES-12 power spectrum occurs at a frequency of approximately 0.979 cycles/day, and the peak on the the Ivan GOES-12 power spectrum occurs at a frequency of 1.0 cycles/day.}
	\label{fig:power}
\end{figure}

After the data is prepared, we apply the steps described in Sec.~\ref{sec:method} to each data set.
For the two Hurricane Felix data sets, as they are from the same hurricane, we would expect the results to be similar despite the temporal and spatial resolution differences.
Plotting the calculated maximum persistence over time, we get the time series plotted as solid lines in Fig.~\ref{fig:smoothed}.
The plots show an oscillatory pattern for all three data sets which appears to repeat approximately daily.

To verify the periodicity of the oscillatory pattern, we apply the discrete Fourier transform and calculate the power spectrum for each data set.
Each power spectrum is shown in Fig.~\ref{fig:power}.
Picking the frequencies corresponding to the highest peaks in each power spectrum gives us the frequency corresponding to the most prominent sinusoidal signal in the data.
The maximum peaks in Fig.~\ref{fig:power} give a frequency of 0.976 cycles per day for the Felix GridSat-GOES data set, 0.979 cycles per day for the Felix GOES-12 data set, and 1.0 cycles per day for the Ivan GOES-12 data set.
We use this frequency, $f$, to calculate the periodicity of the cycle by calculating using $1/f$, giving the period of the sinusoid in days per cycle, then multiply by 24 to rescale to hours per cycle.
Doing so gives the result that the cycle is repeating every 24.6 hours for the Felix GridSat-GOES data set, every 24.5 hours for the Felix GOES-12 data set, and 24.0 hours for the Ivan GOES-12 data set.

Using the most prominent frequency for each data set, we calculate the inverse Fourier transform, and plot these reconstructed sinusoids over the original data. %
These sinusoids, plotted as the lighter dashed lines in Fig.~\ref{fig:smoothed}, closely resemble the patterns exhibited by the original maximum persistence versus time plots; therefore, these approximately 24 hour patterns visible in the plots are also detected mathematically, which verifies the claim that our method is detecting a daily cycle in each data set.
Additionally, since for both Hurricane Felix data sets, the plots of maximum persistence against time seem to match and both have similar detected periodicity from the DFT, our method seems robust to the temporal and spatial resolution differences in these two data sets.

\subsection{Choice of threshold}
\label{ssec:threshold}

\begin{figure}[tb]
	\centering
	\includegraphics[width = .22\textwidth]{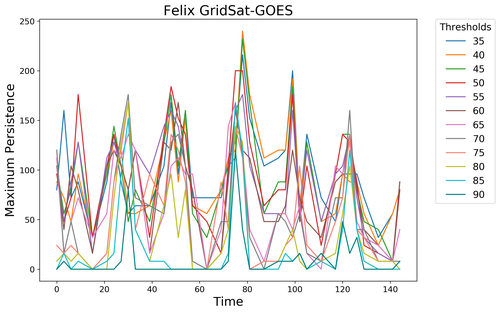} %
	\includegraphics[width = .22\textwidth]{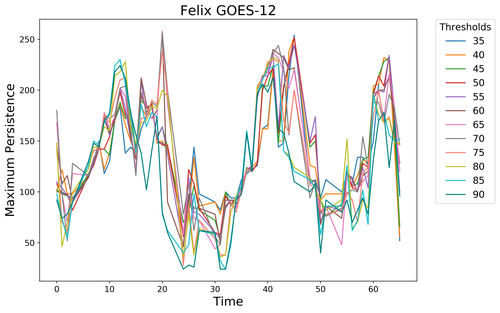} \\
	\includegraphics[width = .22\textwidth]{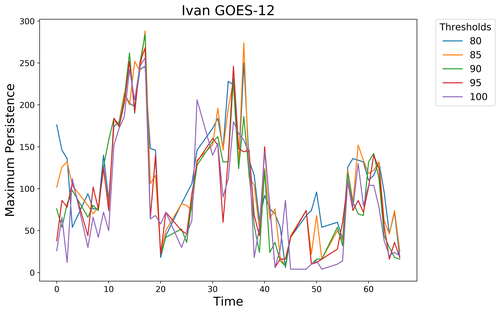}
	\caption{Maximum persistence vs time plot for Hurricane Felix (top row) and Hurricane Ivan (bottom row). Hurricane Felix results are shown for all thresholds $\mu \in \{35,40, \ldots, 90\}$  while Hurricane Ivan results are shown for $\mu \in \{80,85,\ldots,100\}$.}
	\label{fig:many_thresholds}
\end{figure}

The method described involves a choice of threshold, so we used a variety of thresholds, $\mu \in \{25,30,\ldots, 100\}$, to test the sensitivity of our method to the parameter choice.
For both data sets of Hurricane Felix, our method is very robust to the choice of threshold.
In Fig.~\ref{fig:many_thresholds}, the top row are plots that represent maximum persistence versus time for the Hurricane Felix data sets using a variety of thresholds.
There is a clear periodic pattern for both data sets across all the thresholds shown. %
In fact, for all thresholds tested $\mu \in \{35,40 \ldots, 90\}$, the period is consistent at 24.6 hours for the Felix GridSat-GOES data set and 24.5 hours for the Felix GOES-12 data set.
For $\mu < 35$ and $\mu > 90$ the Fourier transform is unable to pick up the daily pattern in the Felix GridSat-GOES data set. %

For Hurricane Ivan, the plot is shown on the bottom row of Fig.~\ref{fig:many_thresholds} for thresholds $\mu \in \{80, 85, \ldots, 100\}$.
For all of the threshold value shown, our method consistently detects a 24.0 hour period.
This is a smaller range of threshold values than those that detect a daily cycle in Hurricane Felix, but for thresholds $\mu \in \{80,85,90\}$, our method detects a daily cycle in all three data sets.
Thus, the method may require some parameter tuning, but our analysis of these three data sets gives a range of values to start with when testing new data sets.

\subsection{Removal of noise}
\label{ssec:noise}

\begin{figure}[tb]
	\centering
	\includegraphics[width = .23\textwidth]{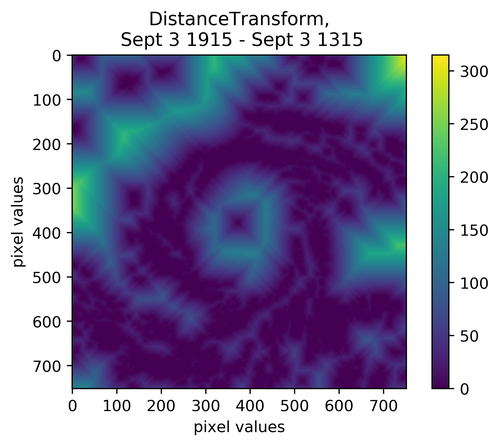}
	\includegraphics[width = .23\textwidth]{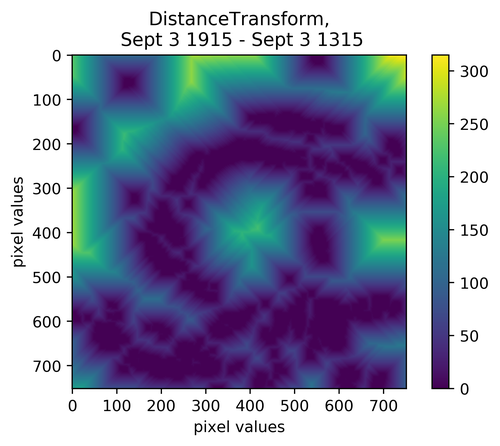}
	\caption{A example comparison of distance transform images with and without opening for the Felix GOES-12 data set.
  At left, the noise in the center of the hurricane causes the distance transform to fill in.
  At right, performing opening gets rid of the small noisy point, and the distance transform does not get filled in.
  }
	\label{fig:center}
\end{figure}

\begin{figure}[tb]
	\centering
	\includegraphics[width = .37\textwidth]{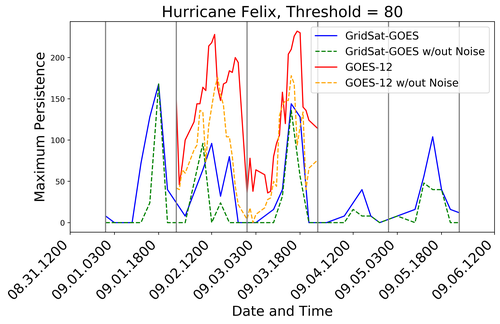} \\
	\includegraphics[width = .37\textwidth]{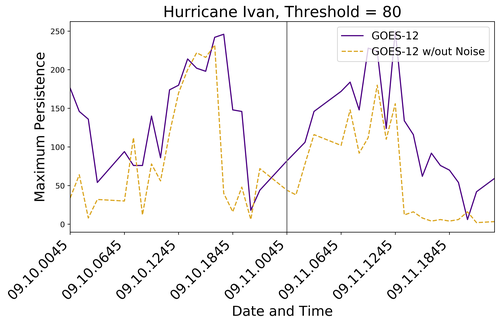}
	\caption{Maximum persistence plotted over time for all data sets using threshold $\mu=80$ in addition to the versions using opening to remove noise. Gray vertical lines separate days according to UTC.}
	\label{fig:MaxPers_open}
\end{figure}

While the above method detects a daily cycle, there are some instances where the six-hour differencing introduces noise because of varying behavior in the center of the hurricane.
The left image of in Fig.~\ref{fig:center} shows an example of how this noise can appear in the distance transform for the Felix GOES-12 data set.
A small area of pixels above the threshold cause the distance transform to fill in the center of the circular region, thus potentially changing the value of maximum persistence.
Therefore, before applying the distance transform, we use a method from mathematical morphology (\cite{serra1984image}) called opening to de-noise the image and see how this impacts the detected periodicity.

Opening is the combination of two tools from mathematical morphology, erosion and dilation.
Both involve a kernel moving through a binary image.
In erosion, a pixel in the original image will remain a 1 only if all pixels under the kernel are 1's, otherwise it becomes a 0.
Dilation is the opposite of erosion,
A kernel moves through the binary image and a pixel is assigned a 1 if at least one pixel under the kernel is a 1, otherwise it is assigned a 0.
Opening is erosion followed by dilation, which will remove noise and rebuild the area around the boundary.

We apply opening to the binary thresholded image using a $8\times 8$ pixel kernel for the GOES-12 data sets and a $2\times 2$ pixel kernel for the GridSat-GOES data set to remove noise such as these center pixels.
Note, the difference in size of the kernel is due to the differences in spatial resolution between the two data sets.
We use the python module \texttt{cv2} for these computations.
Opening is specifically implemented using the function \texttt{cv2.morphologyEx} using \texttt{cv2.MORPH\_OPEN} as the second input.
The right image in Fig.~\ref{fig:center} show the result when opening is used on the thresholded matrix and then the distance transform is applied.
Since the distance transform is no longer filled in, the opening process has removed the noisy pixels causing the issue.

Using this extra step in the method, we recalculate maximum persistence for all times and compute the estimated period of the new maximum persistence values using Fourier transforms. %
Figure~\ref{fig:MaxPers_open} shows maximum persistence plotted versus time for using our original method, and the method including the additional opening step.
While the new maximum persistence values vary a little from the originals, the general oscillatory behavior seems similar.
For both the Felix and Ivan GOES-12 data sets, the Fourier transform still detects a 24.5 and 24.0 hour cycle respectively.
Thus the presence of noise in these data sets is not impacting the results.
However, for the Felix GridSat-GOES data set, the Fourier transform now detects a 15.375 hour cycle, likely due to the difference in spatial resolution.
The GOES-12 data has higher spatial resolution, so applying opening to remove noise does not impact the global circular structure.
The GridSat-GOES data has lower spatial resolution, and is therefore more sensitive to noise in the image.
Thus, our method is more reliable when applied to higher spatial resolution data, and should be used with caution on lower quality data.

%% file: sec-Discussion.tex
\section{Discussion}

This paper presents a novel method for detecting and analyzing the diurnal cycle of tropical cyclones using methods from TDA.
Current state of the art TC diurnal cycle measurement in the satellite imagery is mostly qualitative; our method provides a mathematically advanced method for automatic detection and measurement.
While our method involves a choice of a parameter for the threshold, we present evidence that a range of threshold values yield the same results.

Here, we show that using two sets of GOES satellite data for Hurricane Felix and one set of data for Hurricane Ivan, our method is able to detect almost identical patterns across all three data sets.
Our method performs more consistently and robustly on higher spatial resolution data sets, represented by the GOES-12 data for both Hurricanes Felix and Ivan.
While the method does detect a daily cycle in the lower resolution Felix GridSat-GOES data, when the images are blurred to remove noise, the cycle is no longer detected by the Fourier transform.

This is a novel application of methods from TDA and image processing to the TC diurnal cycle.
We believe this method could be used to study additional atmospheric phenomena exhibiting circular structure.
A future direction of this project is to apply this analysis to more TCs, other satellite channels and other atmospheric data to further test our method.

%% file: DiurnalCyclePersistenceFelix_arxiv.bbl
\begin{thebibliography}{10}

\bibitem{Brigham1988}
E.~O. Brigham.
\newblock {\em The Fast Fourier Transform and Its Applications}.
\newblock Prentice-Hall, Inc., Upper Saddle River, NJ, USA, 1988.

\bibitem{Carlsson2008}
G.~Carlsson, T.~Ishkhanov, V.~de~Silva, and A.~Zomorodian.
\newblock On the local behavior of spaces of natural images.
\newblock {\em International Journal of Computer Vision}, 76:1--12, 2008.

\bibitem{Deckard2013}
A.~Deckard, R.~C. Anafi, J.~B. Hogenesch, S.~B. Haase, and J.~Harer.
\newblock Design and analysis of large-scale biological rhythm studies: a
  comparison of algorithms for detecting periodic signals in biological data.
\newblock {\em Bioinformatics}, 29(24):3174--3180, 2013.

\bibitem{Ditchek2019a}
S.~D. Ditchek, K.~L. Corbosiero, R.~G. Fovell, and J.~Molinari.
\newblock Electrically active tropical cyclone diurnal pulses in the atlantic
  basin.
\newblock {\em Monthly Weather Review}, 147(10):3595--3607, 2019.

\bibitem{Ditchek2019b}
S.~D. Ditchek, J.~Molinari, K.~L. Corbosiero, and R.~G. Fovell.
\newblock An objective climatology of tropical cyclone diurnal pulses in the
  atlantic basin.
\newblock {\em Monthly Weather Review}, 147(2):591--605, 2019.

\bibitem{dunion2019tropical}
J.~P. Dunion, C.~D. Thorncroft, and D.~S. Nolan.
\newblock Tropical cyclone diurnal cycle signals in a hurricane nature run.
\newblock {\em Monthly Weather Review}, 147(1):363--388, 2019.

\bibitem{Dunion2014}
J.~P. Dunion, C.~D. Thorncroft, and C.~S. Velden.
\newblock The tropical cyclone diurnal cycle of mature hurricanes.
\newblock {\em Monthly Weather Review}, 142(10):3900--3919, 2014.

\bibitem{Edelsbrunner2010}
H.~Edelsbrunner and J.~Harer.
\newblock {\em Computational Topology: An Introduction}.
\newblock American Mathematical Society, 2010.

\bibitem{Giusti2015}
C.~Giusti, E.~Pastalkova, C.~Curto, and V.~Itskov.
\newblock Clique topology reveals intrinsic geometric structure in neural
  correlations.
\newblock {\em Proceedings of the National Academy of Sciences}, 2015.

\bibitem{Hatcher}
A.~Hatcher.
\newblock {\em Algebraic Topology}.
\newblock Cambridge University Press, 2002.

\bibitem{KaczynskiMischaikowMrozek2006}
T.~Kaczynski, K.~Mischaikow, and M.~Mrozek.
\newblock {\em Computational homology}, volume 157.
\newblock Springer Science \& Business Media, 2006.

\bibitem{Khasawneh2015}
F.~A. Khasawneh and E.~Munch.
\newblock Chatter detection in turning using persistent homology.
\newblock {\em Mechanical Systems and Signal Processing}, 70-71:527--541, 2016.

\bibitem{Knapp2018}
K.~R. Knapp and S.~L. Wilkins.
\newblock Gridded satellite~({GridSat}) {GOES} and {CONUS} data.
\newblock {\em Earth System Science Data}, 10(3):1417--1425, aug 2018.

\bibitem{kossin2002daily}
J.~P. Kossin.
\newblock Daily hurricane variability inferred from goes infrared imagery.
\newblock {\em Monthly Weather Review}, 130(9):2260--2270, 2002.

\bibitem{lazzara1999man}
M.~A. Lazzara, J.~M. Benson, R.~J. Fox, D.~J. Laitsch, J.~P. Rueden, D.~A.
  Santek, D.~M. Wade, T.~M. Whittaker, and J.~Young.
\newblock The man computer interactive data access system: 25 years of
  interactive processing.
\newblock {\em Bulletin of the American Meteorological Society},
  80(2):271--284, 1999.

\bibitem{leppert2016tropical}
K.~D. Leppert and D.~J. Cecil.
\newblock Tropical cyclone diurnal cycle as observed by trmm.
\newblock {\em Monthly weather review}, 144(8):2793--2808, 2016.

\bibitem{Mischaikow2013}
K.~Mischaikow and V.~Nanda.
\newblock Morse theory for filtrations and efficient computation of persistent
  homology.
\newblock {\em Discrete \& Computational Geometry}, 50(2):330--353, 2013.

\bibitem{Munch2017}
E.~Munch.
\newblock A user's guide to topological data analysis.
\newblock {\em Journal of Learning Analytics}, 4(2), 2017.

\bibitem{Perseus}
V.~Nanda.
\newblock Perseus.
\newblock \url{http://www.sas.upenn.edu/~vnanda/perseus/}, July 2015.

\bibitem{navarro2016idealized}
E.~L. Navarro and G.~J. Hakim.
\newblock Idealized numerical modeling of the diurnal cycle of tropical
  cyclones.
\newblock {\em Journal of the Atmospheric Sciences}, 73(10):4189--4201, 2016.

\bibitem{o2017accessible}
M.~E. O'Neill, D.~Perez-Betancourt, and A.~A. Wing.
\newblock Accessible environments for diurnal-period waves in simulated
  tropical cyclones.
\newblock {\em Journal of the Atmospheric Sciences}, 74(8):2489--2502, 2017.

\bibitem{Perea2015}
J.~A. Perea and J.~Harer.
\newblock Sliding windows and persistence: An application of topological
  methods to signal analysis.
\newblock {\em Foundations of Computational Mathematics}, pages 1--40, 2015.

\bibitem{Robins2016}
V.~Robins, M.~Saadatfar, O.~Delgado-Friedrichs, and A.~P. Sheppard.
\newblock Percolating length scales from topological persistence analysis of
  micro-{CT} images of porous materials.
\newblock {\em Water Resources Research}, 52(1):315--329, jan 2016.

\bibitem{RobinsWoodSheppard2011}
V.~Robins, P.~J. Wood, and A.~P. Sheppard.
\newblock Theory and algorithms for constructing discrete morse complexes from
  grayscale digital images.
\newblock {\em IEEE Transactions on Pattern Analysis and Machine Intelligence},
  33(8):1646--1658, Aug. 2011.

\bibitem{Rosenfeld1968_distancetransform}
A.~Rosenfeld and J.~Pfaltz.
\newblock Distance functions on digital pictures.
\newblock {\em Pattern Recognition}, 1(1):33 -- 61, 1968.

\bibitem{Rosenfeld1966_distancetransform}
A.~Rosenfeld and J.~L. Pfaltz.
\newblock Sequential operations in digital picture processing.
\newblock {\em J. ACM}, 13(4):471--494, Oct. 1966.

\bibitem{Saadatfar2017}
M.~Saadatfar, H.~Takeuchi, V.~Robins, N.~Francois, and Y.~Hiraoka.
\newblock Pore configuration landscape of granular crystallization.
\newblock {\em Nature Communications}, 8(15082), 2017.

\bibitem{serra1984image}
J.~Serra.
\newblock {\em Image Analysis and Mathematical Morphology}.
\newblock Number v. 1 in Image Analysis and Mathematical Morphology. Academic
  Press, 1984.

\bibitem{steranka1984diurnal}
J.~Steranka, E.~B. Rodgers, and R.~C. Gentry.
\newblock The diurnal variation of atlantic ocean tropical cyclone aoud
  distributioninferred from geostationary satellite infrared measurements.
\newblock {\em Monthly weather review}, 112(11):2338--2344, 1984.

\bibitem{tang2016impacts}
X.~Tang and F.~Zhang.
\newblock Impacts of the diurnal radiation cycle on the formation, intensity,
  and structure of hurricane edouard (2014).
\newblock {\em Journal of the Atmospheric Sciences}, 73(7):2871--2892, 2016.

\end{thebibliography}
